\def\year{2018}\relax
\documentclass[letterpaper]{article} 
\usepackage{aaai18}  
\usepackage{times}  
\usepackage{helvet}  
\usepackage{courier}  
\usepackage{url}  
\usepackage{graphicx}  
\usepackage{latexsym}
\usepackage{amsfonts,amssymb}
\usepackage{amsmath}
\usepackage{subfig}
\usepackage{algorithm,algorithmic}
\usepackage{multirow}
\usepackage{xcolor}
\usepackage{tabularx}
\begin{document}
%
\title{VSE-ens: Visual-Semantic Embeddings with Efficient Negative Sampling}
\author{Guibing Guo\thanks{These authors contributed equally to this paper and share the co-first authorship.}\and
Songlin Zhai\footnotemark[1]\and
Fajie Yuan$^{\dagger}$\footnotemark[1]\and
Yuan Liu\and
Xingwei Wang\\
Northeastern University, China \ $^{\dagger}$University of Glasgow, UK\\
\{guogb,liuyuan,wangxw\}@swc.neu.edu.cn, 1771061@stu.neu.edu.cn, f.yuan.1@research.gla.ac.uk
}
\maketitle

\begin{abstract}

Jointing visual-semantic embeddings (VSE) have become a research hotpot for the task of image annotation, which suffers from the issue of \emph{semantic gap}, i.e., the gap between images' visual features (low-level) and labels' semantic features (high-level). This issue will be even more challenging if visual features cannot be retrieved from images, that is, when images are only denoted by numerical IDs as given in some real datasets. The typical way of existing VSE methods is to perform a uniform sampling method for negative examples that violate the ranking order against positive examples, which requires a time-consuming search in the whole label space. In this paper, we propose a fast adaptive negative sampler that can work well in the settings of no figure pixels available. Our sampling strategy is to choose the negative examples that are most likely to meet the requirements of violation  according to the latent factors of images. In this way, our approach can linearly scale up to large datasets. The experiments demonstrate that our approach converges 5.02x faster than the state-of-the-art approaches on OpenImages, 2.5x on IAPR-TCI2 and 2.06x on NUS-WIDE datasets, as well as better ranking accuracy across datasets. 

\end{abstract}
\section{Introduction}
Automatic image annotation is an important task to index and search images of interest from the overwhelming volume of images derived from digital devices. It aims to select a small set of appropriate labels or keywords (i.e., annotations) from a given dictionary that can help describe the content of a target image. However, it is not trivial to handle the differences between low-level visual features of images and high-level semantic features of annotations, which has been well recognized as the problem of \emph{semantic gap}. This issue becomes even more challenging if no visual features can be drawn from figure pixels, that is, when images are only represented by numerical IDs rather than pixel values. This problem  setting can be observed in some real datasets, which is the target scenario of this paper. 

A promising way to resolve this issue is to jointly embed images and annotations into the same latent feature space, a.k.a. visual-semantic embeddings (VSE)~\cite{weston2011wsabie,DBLP:journals/corr/FaghriFKF17}. Since both images and annotations are represented by the same set of latent features, their semantic differences can be converged and computed in the same space. Existing VSE methods are derived in the form of pairwise learning approaches. That is, for each image, a set of pair-wised (positive, negative) annotations will be retrieved to learn a proper pattern to represent the image. Due to the large volume of negative candidates, it is necessary to take sampling strategies in order to form balanced training data. The most frequently adopted strategy, e.g. in \cite{weston2011wsabie}, is to repeatedly sample negative labels from the dictionary that violates the ranking order against positive examples. However, the whole annotation space may need to be traversed until a \emph{good} negative example is found. In a word, it is time-consuming and thus cannot be applied to large-scale datasets.

In this paper, we propose a fast adaptive negative sampler for the task of image annotation based on joint visual-semantic embeddings (VSE). It is able to well function in the problem settings of no figure pixels available. Instead of traversing the whole annotation set to get good negative examples, we selectively choose those labels that are most likely to meet the requirements of violation according to the latent factors of images and annotations. Specifically, our proposed sampler adopts a rank-invariant transformation to dynamically select the required high-ranked negative labels without conducting the inner product operations of the embedding vectors. In this way, the running time of negative sampling can be dramatically reduced. We conduct extensive experiments on three real datasets (OpenImages\footnote{\footnotesize \url{https://github.com/openimages/dataset}}, IAPR-TCI2\footnote{\footnotesize \url{http://www.imageclef.org/photodata}}, NUS-WIDE\footnote{\small \url{http://lms.comp.nus.edu.sg/research/NUS-WIDE.htm}}) to demonstrate the efficiency of our approach. The results show that our method is 5.02 times faster than other state-of-the-art approaches on OpenImages, around 2.5 times on IAPR-TCI2 and 2.06 times on NUS-WIDE at no expense of ranking accuracy.

Our main contributions of this paper are given as follows.
\begin{itemize}
	\item We propose a fast adaptive sampler to select good negative examples for the task of image annotation. It adopts a rank-invariant transformation  to dynamically choose highly ranked negative labels, whereby the time complexity can be greatly reduced.
    \item We provide the corresponding proof to show that the proposed sampling is theoretically equivalent  with the inner product based negative sampling, and thus ensure comparable and even better performance in ranking accuracy.
    \item We conduct a series of experiments on three real image-annotation datasets. The results further confirm that our approach performs much faster than other counterparts in terms of both training time and ranking accuracy.
\end{itemize}

\section{Preliminary}
In what follows, we first introduce the visual-semantic embeddings. Then we summarize the typical negative sampling algorithm used in WARP \cite{weston2011wsabie} and point out its inefficiency issue.

\subsection{Visual-Semantic Embedding}
Following WARP, we start with a representation of images $\emph{i}\in \mathbb{R}^{d}$ and a representation of annotations $\emph{a} \in \emph{A} = \{\emph{a}_1 , \emph{a}_2, ..., \emph{a}_m \}$ to indicate an annotation of a dictionary.  Let $C = \left\{ (i_m,a_m)\right\}^M_{m=1}$ denote a training set of image-annotation pairs. We refer to $(i_m,a_p)$ as positive pairs while $(i_m,a_{n})$ as negative pairs\footnote{That is, the annotation $a_n$ is not labeled on image $i_m$.}. $s_i(a)$ is an inner product function that calculates a relevance score of an annotation $a$ for a given image $i$ under the VSE space. $V \in \mathbb{R}^{(d+|A|) \times  k  }$ denotes the embedding matrix of both images and annotations, where $\mathbb{R}^{d\times k}$ corresponds to image embedding matrix while $\mathbb{R}^{|A|\times k}$ corresponds to annotation embedding matrix and $k$ is the embedding dimension. Meanwhile, we have the function $W_I (i)$ that maps the image feature space $\mathbb{R}^{d}$ to the embedding space $\mathbb{R}^{k}$,
and  $W_A(a)$ jointly maps  annotation space from $\mathbb{R}^{|A|}$ to $\mathbb{R}^{ k }$. Assuming a linear map is chosen for $W_I (i)$ and $W_A(a)$, we can have $W_I (i)$ = $v_i$ and $W_A(a)$=$v_a$, where $v_i$ and $v_a$ are the $i$-th and $a$-th row of $V$.

Hence, we consider the scoring function as follows:

\begin{equation}
	s_i(a)={W_I (i)}^T \cdot W_A(a)=\sum_{f=1}^k  v_{if} v_{af}
    \label{eq:score}
\end{equation}
where $f$ is the embedding factor and the magnitude of $s_i(a)$ denotes the relevance between $a$ and $i$. The goal of VSE is to score the positive pairs higher than the negative
pairs.  With this in mind, we consider the task of image annotation as a standard ranking problem. 

\subsection{The WARP Model}
WARP~\cite{weston2011wsabie} is known as a classical optimization approach for joint visual-semantic embeddings, where a weighted approximate-rank pairwise loss is applied.  The loss function is generally defined by Eq.~\ref{eq:err}, which enables the optimization of precision at $N$ by stochastic gradient descent (SGD).
\begin{equation}
	\overline{err}(s_i(a),a_p)\! = \!\!\sum_{p \neq n} L(rank(s_i(a_p))) \frac{| 1 - s_i(a_p) + s_i(a_n) |_{+}}{rank(s_i(a_p))}
    \label{eq:err}
\end{equation}
where $rank(s_i(a_p))$ is a function to measure how many negative annotations are `wrongly' ranked higher than the positive ones $a_p$, given by:
\begin{equation}
	rank\big(s_i(a_p)\big) = \sum_{p \neq n} I(1 + s_{i}(a_n) > s_i(a_p))
    \label{eq:rank}
\end{equation}
where $I(\cdot)$ is an indicator function. The function $L(\cdot)$ transforms the rank into a loss, defined by:
\[
	L(k) = \sum_{j=1}^k \xi_j, ( \xi_1 \geq \xi_2 \geq ... \geq 0)
\]
where $\xi_j$ defines the importance of relative position of the positive example in the ranked list, e.g., $\xi_j = \frac{1}{|A| - 1}$  is used to optimize the mean rank.

The overall risk that needs to minimize is given by:
\[
	Risk(s) = \int \overline{err}(s_i(a),a_p) dP(i,a_p)
\]
where $P$ indicates the probability distribution of positive image-annotation pair $(i,a_p)$, which is a uniform distribution in WARP.

\begin{figure*}[tb]
	\centering
	\begin{minipage}[c]{0.33\textwidth}
		\subfloat[OpenImages]{\includegraphics[width=60mm]{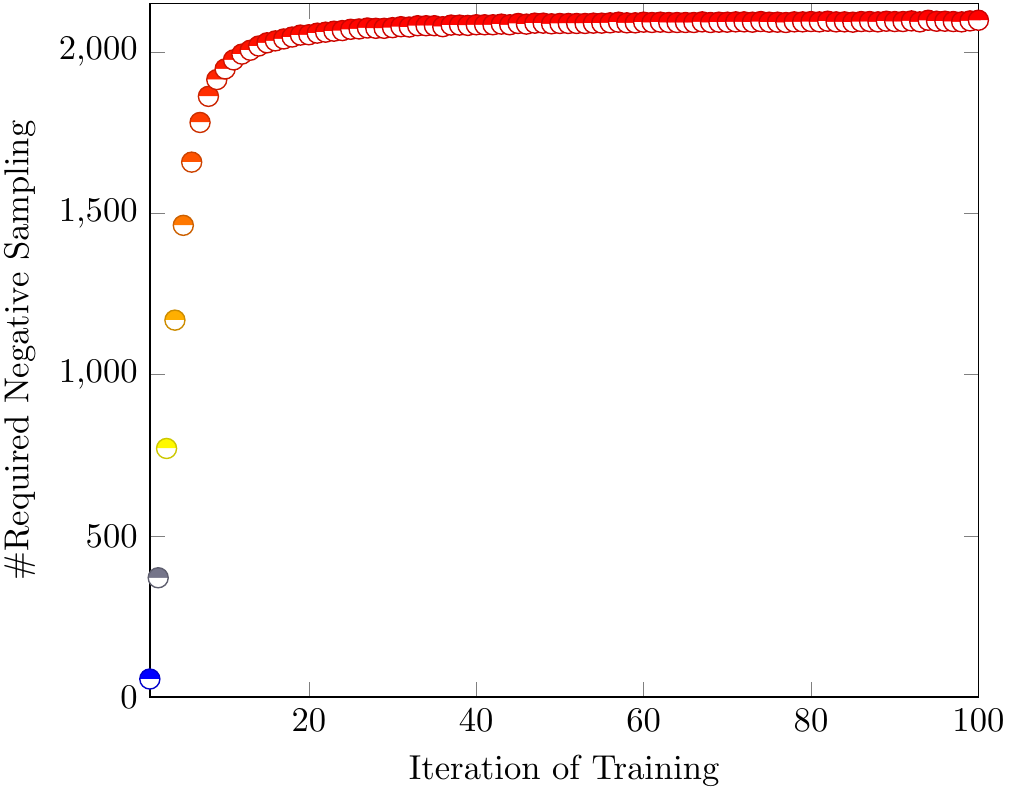}}
	\end{minipage}
	\begin{minipage}[c]{0.33\textwidth}
		\subfloat[NUS-WIDE]{\includegraphics[width=60mm]{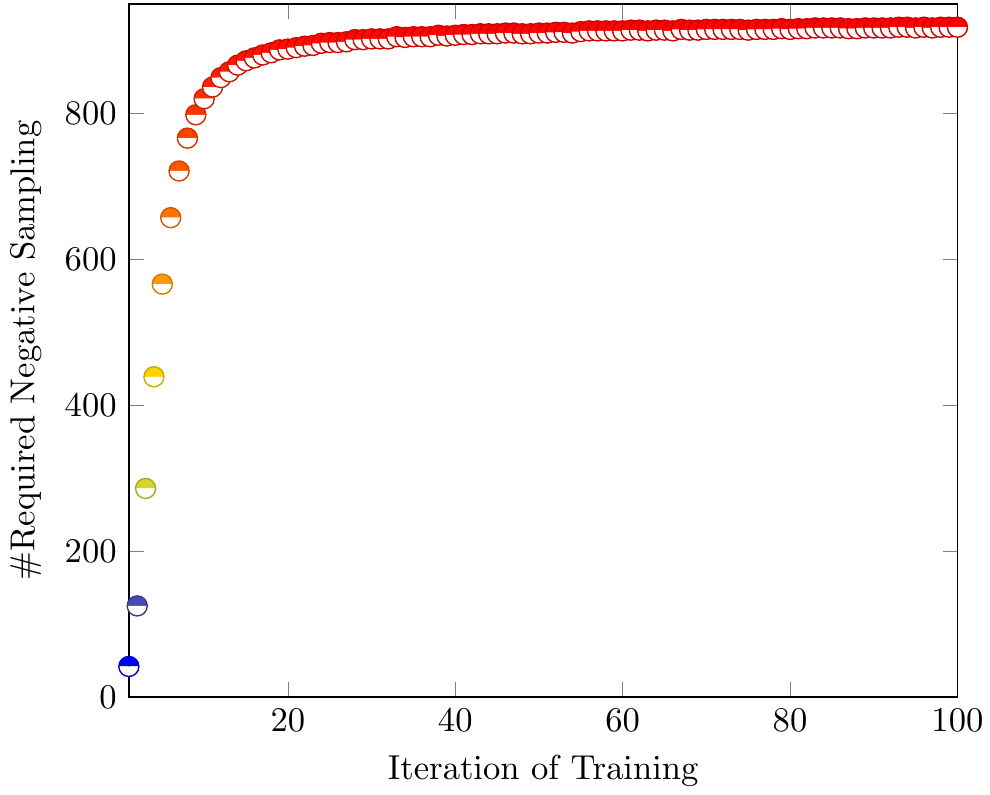}}
	\end{minipage}
	\begin{minipage}[c]{0.33\textwidth}
		\subfloat[IAPR-TC12]{\includegraphics[width=60mm]{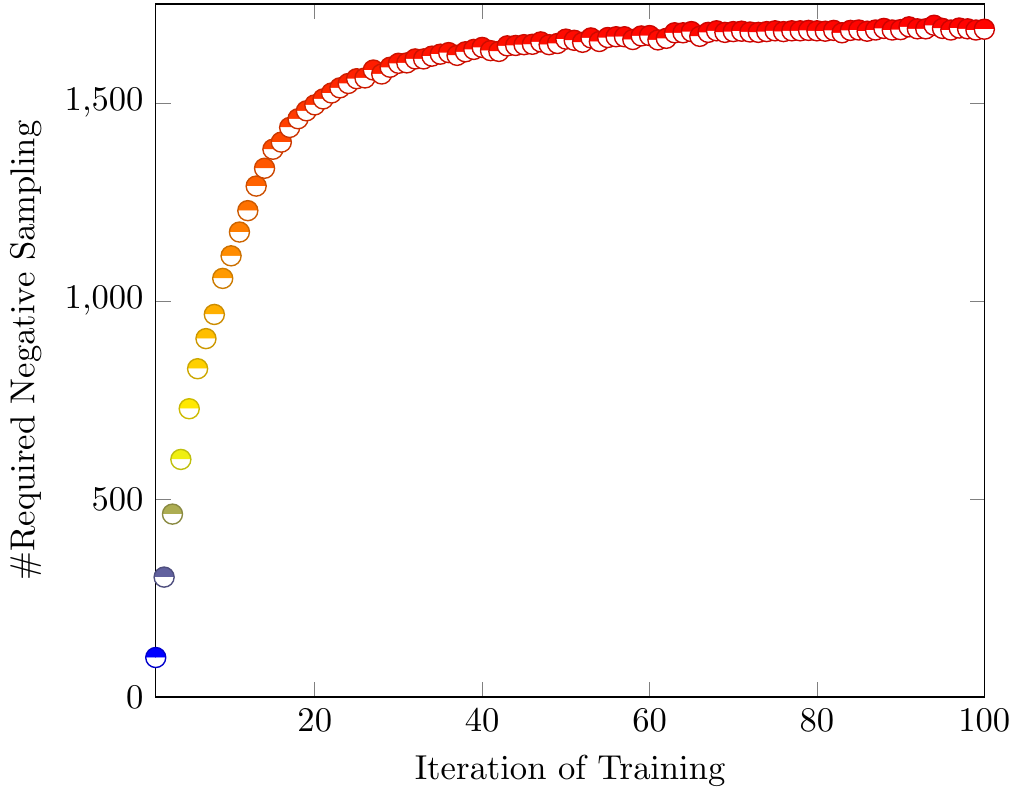}}
	\end{minipage}
	\caption{The number of required negative samples of the WARP model as the SGD iterations increase. }
	\label{fig:warp-freq}
	\vspace{-5mm}
\end{figure*}

\subsection{Negative Sampling}
An unbiased estimator of the above risk can be obtained by stochastically sampling in the following steps:

\begin{enumerate}
	\item Sample a positive pair $(i,a_p)$ according to $P(i,a_p)$.
	\item Repeatedly sample a required annotation $a_n$ such that:
          \begin{equation}
              1 + s_i(a_n) > s_i(a_p)
              \label{eq:requirement}
          \end{equation}
\end{enumerate}

This chosen triplet $(i, a_p, a_n)$ contributes to the total risk:
\[
\overline{err}(s_i(a),a_p, a_n)
= L(rank(s_i(a_p))) | 1 - s_i(a_p) + s_i(a_n) |_{+}
\]
The sampling strategy in step 2 generally implies that the learning algorithm concentrates merely on the negative annotation with a higher score, i.e., $s_i(a_n) > s_i(a_p)-1$.
The idea is intuitively correct since  negative examples with higher scores are more likely to be ranked higher than positive ones, and thus results in a larger loss \cite{yuan2016lambdafm,yuan2017boostfm}. Hence, as long as the learning algorithm can distinguish these higher scored negative examples, the loss is supposed to be diminished to a large extent.

\subsection{Efficiency Issue of the WARP Sampler}
Even though WARP has been successfully applied in various VSE scenarios, in the following it is shown that the computational cost of WARP sampling is expensive, in particular when it has been trained after several iterations.
As depicted in step 2, a repeated sampling procedure has to be performed such that a required negative example can be observed. The computational complexity of scoring a negative pair in Eq.~\ref{eq:requirement} is in $O(k)$. In the beginning, since the model is not well trained, it is easier to find a violated negative example that has a higher score than the positive one, which leads to a complexity of $O(Tk)$, where $T$ is the average sampling trials. However, after several training iterations, most positive pairs are likely to have higher scores than negative ones, and thus $T$ becomes much bigger, with the complexity up to  $O(|A|k)$, where $|A|$ is the size of the whole annotation set. For each SGD update, the sampler may have to iterate all negative examples in the whole annotation collection, which is computationally prohibitive for large-scale datasets.

\subsection{Experimentation on the Efficiency Issue}
According to Eq.~\ref{eq:requirement}, the WARP sampler always attempts to find the violating annotation for a given image annotation pair. Along with the convergence of WARP training, most annotations have met the demand ($ s_i(a_p) > 1 + s_i(a_n)$), and thus it will take longer time per iteration to find the expected violation annotation. To verify our analysis, we conduct experiments on three datasets to count the number of required negative sampling in the WARP model, as illustrated in Figure~\ref{fig:warp-freq}. We defer the detailed description of datasets to the evaluation section. 

Specifically, the number of required negative sampling increases very drastically before the 13th iteration, which takes over 2,000 repeated sampling until finding an appropriate example. After that, the number stays high at about 2,100 on the OpenImages dataset. For the NUS-WIDE dataset, before the 15th iteration, the required sampling grows rapidly up to 870 and then stable at around 900. Analogously, the number of negative sampling quickly increases at the beginning stage and then keeps stable at a high value around 1,600 on the IAPR-TC12 dataset.

To sum up, the WARP sampler will become slower and slower as the SGD update iterations accumulate. Hence, we aim to resolve this issue in this paper by proposing a novel and efficient negative sampling method for the VSE field.

\section{Fast Sampling Algorithm}
Actually, suchlike sampler has been adopted not only in the visual-semantic embedding task but also in many other fields.
For example, in \cite{weston2012latent},   \cite{hsiao2014social} and \cite{li2015rank},
they successfully applied the WARP loss function for the collaborative retrieval/filtering tasks and achieved state-of-the-art results. Inspired by this, 
we\footnote{\scriptsize Though our previous AAAI version \cite{guo2018vse} has cited \cite{rendle2014improving} for the purpose of relevance, we'd like to make a further clarification here to avoid potential misunderstanding.} attempt to adapt the sampling strategy in \cite{rendle2014improving}  to solve the above-mentioned inefficient issue of the original sampler
in our visual-semantic embedding task,  which is a different research domain from \cite{rendle2014improving}\footnote{\scriptsize Also note the sampling idea in \cite{rendle2014improving} was claimed  to improve only BPR-style \cite{rendle2009bpr} learners which are based on the negative log-likelihood loss, whereas WARP is actually a different  (see \cite{gao2015consistency}) one, which has the non-smooth \& non-differentiable  loss with different gradients. }. In this work, we aim to study the effectiveness of this alternative sampling strategy in speeding up the sampling process and improving the performance boundaries.
 \vspace{-0.05in}
\subsection{Naive Sampling}
As aforementioned, the major computational cost of WARP is caused by the repeated inner product operations in Eq.~\ref{eq:requirement}, which have a complexity of $O(k)$ in each operation.

In the following, an alternative sampler with fast sampling rate is derived which has the same intuition with the negative sampler in WARP ---
considering a negative example $a_n$ for a given positive pair $(i, a_p)$, the higher score $(i, a_n)$ has, the more chance $a_n$ should be sampled. Instead of using the notion of a large score, we opt to formalize a small predicted rank $\hat{r}_i(a_n)$ because the largeness of scores is relative to other examples, but the ranks are absolute values. This allows us to formulate a sampling distribution, e.g., an exponential distribution\footnote{\scriptsize In practice, the distribution can be replaced with other analytic distributions, such as geometric and linear distributions.}, based on annotation ranks such that higher ranked annotations have larger chance to be selected. 
\begin{equation}
	p_i(a_n) \varpropto exp(-\hat{r}_i(a_n)) / \lambda
    \label{eq:eq8}
\end{equation}

Hence, a naive sampling algorithm can be easily implemented by: 
\begin{enumerate}
\item Adopt the  exponential distribution to sample a rank $r$.
\item Return the annotation $a_n$ currently at the ranking position of $r$, i.e. find $a_n$
with $\hat{r}_i(a_n)=r$ or $j=\hat{r}^{-1}_i(a_n)$.
\end{enumerate}

However, it should be noted that this trivial sampling method has to  compute $s_i(a_n)$ for all $a_n$ in $A$, and then sort them by their scores and return the  annotation at place $r$. This algorithm has a complexity of $O(|A|k+|A|log|A|)$ for each SGD learning, which is clearly infeasible in practice.

Motivated by this, we will introduce a more efficient sampling method in the following. The basic idea of our proposed sampler is to formalize Eq.~\ref{eq:eq8} as a mixture of ranking distributions over normalized embedding factors such that the expensive inner production operation can be got around. The mixture probability is derived from a normalized version of the inner product operation in Eq.~\ref{eq:score}.
\subsection{Rank-Invariant  Transformation}
According to Eq.~\ref{eq:score}, a transformation $s^{*}_i(a)$ of $s_i(a)$ can be defined by:
\begin{equation}
	s^{*}_i(a) := \sum^{k}_{f = 1}p(f|i)sgn(v_{i,f})v^{*}_{a,f}
    \label{eq:s}
\end{equation}
where $p(f|i)$ is the probability function that denotes the importance of the latent dimension $f$ for  the image $i$ --- the larger $|v_{i,f}|$ and  $\sigma_{f}$, the more important dimension $f$:

\begin{equation}
	p(f|i) := |v_{i,f}| \cdot \sigma_{f}
    \label{eq:eq10}
\end{equation}
and $v^{*}_{a,f}$ is a standardized label factor if we assume $v_{a,f}$ corresponds to the normal distribution:
\[
	v^{*}_{a,f} = \frac{v_{a,f} - \mu_f}{\sigma_f}
\]
where $\mu_f$ and $\sigma_f$ are the empirical mean and standard deviation  over all labels' factors, given by:
\begin{equation}
	\mu_f = E(v,f), \quad \sigma^{2}_{f} = Var(v,f)
\end{equation}

\noindent 
The main idea is that \emph{\textbf{the ranking $\hat{r}^*$  derived from scoring $s^*$ has the same effect as the ranking $\hat{r}$ from $s$.}}\\

We can prove this as follows:
\[
\begin{aligned}
s_i(a) & = \sum_{f=1}^{k} \ v_{i,f} v_{a,f} \\
& =  \sum_{f=1}^{k} \ | v_{i,f}| \ sgn(v_{i,f}) \ (\sigma_f v^*_{a,f} + \mu_f)  \\
& = \sum_{f=1}^{k} \ | v_{i,f}| \ sgn(v_{i,f}) \ \sigma_f v^*_{a,f} + | v_{i,f}| \ sgn(v_{i,f}) \ \mu_f \\
& = s^*_i(a) + \sum_{f=1}^{k}| v_{i,f}| \ sgn(v_{i,f}) \ \mu_f \\
\end{aligned}
\]
Note that the second term $\sum_{f=1}^{k}| v_{i,f}| \ sgn(v_{i,f})\mu_f$ is independent of label $a$, whereby we can treat it as a constant value. In other words, the ranks generated by $s^*_i(a)$ will be equal with those generated by $s_i(a)$, i.e., $\hat{r} = \hat{r}^*$.

\paragraph{Sampler Function.}
Since the ranks generated by $s_i(a)$ can also work with $s^*_i(a)$, we can define our sampler function according to this characteristic. The representation of $s^*_i(a)$ in Eq.~\ref{eq:s} indicates that the larger $p(f|i)$ is, the more important dimension $f$ is for the specific image $i$. We can define the sampling distribution as follows:
\[
	p(a|i) := \sum_{f=1}^k p(f|i) \ p(a|i,f) 
\]
As $v^*_{.,f}$ has been standardized,  we may define $p(a|i,f)$ in the same manner as Eq.~\ref{eq:eq8}:
\[
	p(a|i,f) \varpropto exp(-\hat{r}^*(a|i,f) / \lambda)
\]
Following Eq.~\ref{eq:s}, the scoring function under the given image $i$ and dimension $f$ can be defined by:
\[
	s^*(a|i,f) := sgn(v_{i,f}) \ v^*_{a,f}
\]
According to the inference aforementioned, the above sampler function can be written as follows:
\begin{equation}
	s(a|i,f) := sgn(v_{i,f}) \ v_{a,f}
    \label{eq:eq16}
\end{equation}
From our sampler function, we can observe an intuitive relation between $s(a|i,f)$ and $\hat{r}(a|i,f)$: the label on rank $r$ has the $r$-th largest factor $v_{a,f}$, if $sgn(v_{i,f})$ is positive; otherwise it has the $r$-th largest negative factor.

\subsection{Process of Sampling}
According to our sampler function (Eq.~\ref{eq:eq16}), the process of sampling negative labels is elaborated as follows:
\begin{enumerate}
\item Draw a rank $r$ from an exponential distribution, e.g., $p(r) \propto exp(-r/\lambda)$.
\item Draw the embedding dimension $f$ from $p(f|i) \propto |v_i,f|\sigma_f$.
\item Sort labels according to $v_{.,f}$. Due to the rank-invariant property, it is thus equivalent to an inverse ranking function ($\hat{r}^{-1}$).
\item Return the label $a_n$ on position $\hat{r}$ in the sorted list according to the value of $sgn(v_{i,f})$, i.e., $\hat{r}(r|f)$ if $sgn(v_{i, f})=1$, or $\hat{r}(|A|-r+1|f)$ if $sgn(v_{i, f})=-1$.
\end{enumerate}

\begin{algorithm}[tb]
\caption{VSE-ens with fast negative sampling}
\label{algo:sampling}
\begin{algorithmic}[1]
\STATE Randomly initialize $\Theta$, $I$, $A$, $q = 0$
\REPEAT
\STATE $ q \leftarrow q + 1 $
\IF{$q \% |A|log|A| = 0$}
\STATE $\triangleright every \ |A|log|A| \ draws $
\FOR{$f \in {1,...,k}$}
\STATE Compute $r^{-1}(.|f)$
\STATE Compute $\sigma^2_f$ \ and \ $\mu_f$
\ENDFOR
\ENDIF
\STATE Draw $(i,a_p)$ $\propto P(i,a_p)$ 
\STATE Draw \emph{r} from $p(r) \propto exp(-r/\lambda)$
\STATE Draw \emph{f} from $p(f|i) \propto |v_i,f|\sigma_f$
\IF{$sgn(v_{i,f})$ = 1}
\STATE $j = r^{-1}(r|f)$
\ELSE
\STATE $j = r^{-1}(|A| -r + 1|f)$
\ENDIF
\FOR{$\theta \in \Theta$}
\STATE $\theta \leftarrow \theta - \eta \nabla_{\theta} | 1 - s_i(a_p) + s_i(a_n) |_{+}$
\ENDFOR
\UNTIL{convergence}
\STATE \textbf{return} $\Theta$
\end{algorithmic}
\end{algorithm}

In the process, it takes $O(1)$ to perform steps 1 and 4, and only costs $O(k)$ to compute $p(f|i)$ in step 2. However, step 3 is computationally expensive to be performed, 
 since the  factors are sorted in $O(|A| \ log|A|)$.

It will take much time if we have to re-sort the ranks in order to get the negative examples for every dimension $f$. Instead, we opt to further reduce the complexity by pre-computing the $k$ rankings for every $|A| \ log|A|$ iterations. This is because the ordering changes only little and many update steps are necessary to change the pre-computed ranking considerably. As a result, the overall complexity $O(k \ |A| \ log|A|)$ can be allocated by $|A| \ log|A|$ iterations. In other words, the additional complexity is just $O(k)$ for each SGD update.

To sum up, the sampling algorithm takes an amount of $O(k)$ computational time to sample a negative annotation which is the same required cost as a single SGD step. As a result, the proposed sampling and SGD process together will not increase much of computational cost. 

Algorithm~\ref{algo:sampling} sketches the pseudocodes of the improved learning algorithm. To explain, several arguments are taken as input, including the model parameters $\Theta$, the collection of images $I$, the collection of annotations $A$ and a variable $q$. Firstly, we precompute the $r^{-1}(.|f)$, $\sigma^2_f$ \ and \ $\mu_f$ with a constant time (line 7 and line 8). Then, we sample an image-annotation pair (line 11) and get the position of this annotation in the annotation embedding space (line 12). Next, we choose a factor in the annotation embedding (line 13) space according to $p(f|i)$ and get another annotation according the value of $sgn(v_{i,f})$ (line 14 - line 17). Finally, we adopt the popular Stochastic Gradient Descent (SGD) algorithm to train our model and update the model parameters $\Theta$ (line 19 and line 20) until convergence.

\begin{figure}[tb]
\centering
\begin{minipage}[c]{\textwidth}
	\includegraphics[width=85mm]{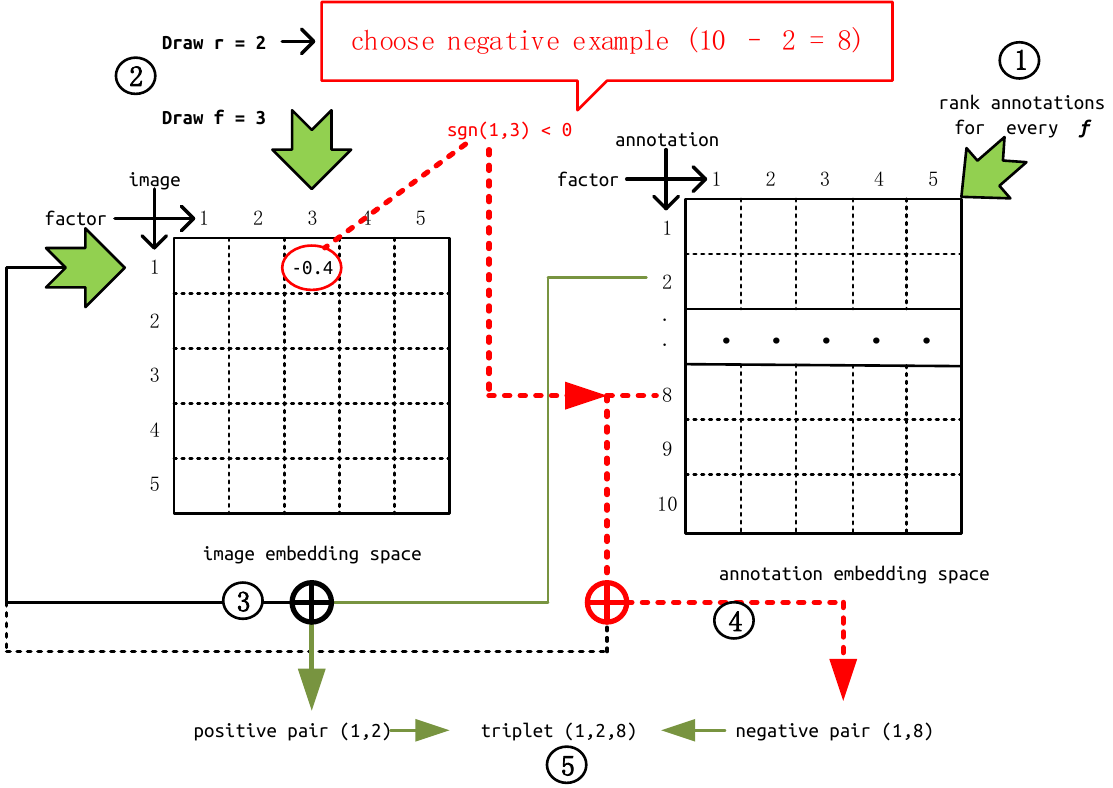}
\end{minipage}
\caption{Example of our adaptive negative sampling}
\label{fig:example}
\end{figure}

\medskip
\noindent\textbf{Example of Negative Sampling:} As shown in Figure~\ref{fig:example}, suppose we have 5 images with 10 annotations in the training datasets and set the number of embedding factor as 5. 
Following Algorithm~\ref{algo:sampling}, our model will rank these annotations according to $v_{a,f}$ for each dimension $f$, and compute the value of $\sigma_f$ and $\mu_f$ at the first iteration. Then, it will randomly choose a positive image-annotation pair, e.g. the 1st image and the 2nd annotation, denoted as $(1, 2)$. After this, the negative sampler will sample a rank $r$, e.g. $r=2$ according to the designed distribution and a dimension $f$, e.g. $f=3$. Finally, we are able to return the negative example according to $sgn(v_{1,3})$, i.e., choosing the negative annotation from the ranked list with $r=8, f=3$ if $sgn(v_{1,3})<0$,  and $r=2,f=3$ if $sgn(v_{1,3})>0$.

\section{Experiments and Results}
\subsection{Datasets}
Three real datasets are used in our evaluation, namely \emph{OpenImages}, \emph{NUS-WIDE} and \emph{IAPR-TC12}. OpenImages is introduced by~\cite{openimages} and contains 9 million URLs to images that have been annotated with image-level labels. NUS-WIDE \cite{chua2009nus} is collected at the National University of Singapore, and composed of 269,648 images annotated with 81 ground-truth concept labels and more than 5,000 labels. IAPR-TC12 produced by \cite{grubinger2006iapr} has 19,627 images comprised of natural scenes such as sports, people, animals, cities or other contemporary scenes. Each image is annotated with an average of 5.7 labels out of 291 candidates. The statistics of the three datasets are presented in Table~\ref{tab:stats}, where rows `Train' and `Test' indicate the number of image-annotation pairs in the training and test set, respectively.

\begin{table}[tb]
\centering
\caption{The statistics of our datasets}
\label{tab:stats}
\begin{tabular}{|l|c|c|c|}
\hline
\rule{0pt}{2.5ex} \textbf{Feature} & \textbf{OpenImages} & \textbf{NUS-WIDE} & \textbf{IAPR-TC12} \\[0.5ex] \hline
Images &112,247  &269,648   &19,627\\
Lables &6,000      &5108      &291\\
Train  &887,752    &2,018,879   &79,527\\
Test   &112,247    &267,642   &20,000\\
\hline
\end{tabular}
\end{table}

\subsection{Experimental Setup}
We have implemented and compared with the following two strong baselines. 
\begin{itemize} 
\item \textbf{WARP}~\cite{weston2011wsabie} uses a negative sampling based weighting approximation (see Eq.~\ref{eq:rank}) to optimize standard ranking metrics, such as precision.
\item \textbf{Opt-AUC} is to optimize Area Under the ROC Curve (AUC). Logistic loss is used as the smoothed AUC surrogate.
\end{itemize}

We adopt the leave-one-out evaluation protocol. That is, we randomly select an annotation from each image for evaluation, and leave the rest for training. All reported results use the same embedding dimension of k = 100. The hyperparameters for VSE-ens on all three datasets are: learning rate $\eta = 0.01$, $\mbox{regularization}=0.01$,  and variables are initialized by a normal distribution $\mathcal{N}(0,0.01)$. Parameter $\lambda$ for VSE-ens is tuned from 0.001 to 1.0 to find the best value. The learning rate and regularization settings of other models are tuned from 0.001 to 0.1 to search the optimal values. 

\setlength{\tabcolsep}{10pt}
\renewcommand\arraystretch{1.2}
\begin{table*}[tb]
\centering
\caption{The ranking accuracy of comparison methods, where the last line of each dataset `Improve' indicates the improvements our approach achieves relative to WARP.}
\label{tab:accuracy}
\begin{tabular}{|l|l|c|c|c|c|c|c|}
\hline
\rule{0pt}{2.5ex} \textbf{Dataset} & \textbf{Model} & \textbf{Pre@5} & \textbf{Rec@5} & \textbf{Pre@10} & \textbf{Rec@10} & \textbf{MAP} & \textbf{AUC}\\[0.5ex] \hline
Open- & VSE-ens &0.0574 &0.2869 &0.0434 &0.4342 &0.1762 &0.7168\\
Images & WARP &0.0526 &0.2628 &0.0390 &0.3900  &0.1676 &0.6948\\
& Opt-AUC &0.0188 &0.0938 &0.0147 &0.1465  &0.0564 &0.5732\\  \cline{2-8}
& Improve & 9.13\% & 9.17\% & 11.28 \% & 11.33 \% & 5.13 \% & 3.17 \% \\ \hline
\hline
NUS- & VSE-ens &0.0278 &0.1391  &0.0198 &0.1982 &0.0893 &0.5990\\
WIDE & WARP &0.0107 &0.0533 &0.0083 &0.0830  &0.0336 &0.5415\\
& Opt-AUC &0.0035 &0.0177 &0.0028 &0.0279  &0.0113 &0.5139\\ \cline{2-8}
& Improve & 159.81 \% & 160.98 \% & 138.55 \% & 138.80 \% & 165.77 \% & 10.62 \% \\ \hline
\hline
IAPR- & VSE-ens &0.0598 &0.2990  &0.0436 &0.4364 &0.1836 &0.7126\\
TC12 & WARP &0.0595 &0.2976  &0.0428 &0.4278 &0.1796 &0.7086\\
& Opt-AUC &0.0543 &0.2713  &0.0414 &0.4136 &0.1629 &0.7011\\ \cline{2-8}
& Improve & 0.50 \% & 0.47 \% & 1.87 \% & 2.01 \% & 2.23 \% & 0.56 \%  \\ \hline
\end{tabular}
\end{table*}

\subsection{Evaluation Metrics}
We use four widely used ranking metrics to evaluate the performance of all comparison methods. Generally, the higher ranking metrics are, the better performance we get. The first two ranking metrics are precision@N and recall@N (denoted by Pre@N and Rec@N). We set $N=5, 10$ for the ease of comparison in our experiments.
\[
	\mbox{Pre@N} = \frac{TP}{TP + FP} \qquad
	\mbox{Rec@N} = \frac{TP}{TP + TN}
\]
where $TP$ is the number of annotations contained in both the ground truth and the top-N results produced by the algorithm;
$FP$ is the number of annotations in the top-N produced results but not in the ground truth; and $TN$ is the number of annotations contained in ground truth but not in the top-N generated results.

We also report the results in Mean Average Precision (MAP) and Area Under the Curve (AUC), which take into account all the image labels to evaluate the full ranking.
\[
	\mbox{MAP} = \frac{\sum_{q = 1}^Q \mbox{AveP}(q)}{Q}
\]
where $Q$ denotes the sample space and $q$ is an example of $Q$. $AveP = \sum_{k = 1}^n P(k) \triangle r(k)$, where $P$ and $r$ denote the Precision and Recall, respectively.

\[
    \mbox{AUC} = \frac{1}{|D_s|} \sum_{(i,a_p,a_n) \in D_s} \frac{\sigma(\hat{x}_{ipn} > 0)}{|I||A_p||A_n|}
\]
where $D_s$ denotes the set of training triplet pairs; $\sigma(\cdot)$ is a sigmoid function and $\hat{x}_{ipn}={f}_i(a_p)-{f}_i(a_n)$ aims to capture the relationship between positive annotation $a_p$ and negative annotation $a_n$ for image $i$.

\subsection{Comparison in Training Time}
We compare the different models in terms of training time. Specifically, Table~\ref{tab:complexity} summarizes the theoretical time complexity of all the comparison methods by iterating all annotation sets; and Table~\ref{tab:time} shows the specific training time on the OpenImages, NUS-WIDE and IAPR-TC12 datasets. The results show that our approach gains up to 5.02 times improvements in training time compared with other comparison methods in the OpenImages dataset.

\setlength{\tabcolsep}{30pt}
\begin{table}[tb]
\centering
\caption{The theoretical time complexity of all the comparison models in each iteration, where \emph{k} is the size of the embedding space, \emph{T} is the average number of sampling trials for negative sampling.}
\label{tab:complexity}
\begin{tabular}{|l|l|}
\hline
\rule{0pt}{2.5ex} \textbf{Model} & \textbf{Time Complexity} \\[0.5ex] \hline
VSE-ens & $ O (2k)$ \\
WARP & $O(Tk)$ \\
Opt-AUC & $O(k)$ \\ \hline
\end{tabular}
\end{table}

\setlength{\tabcolsep}{4.5pt}
\begin{table}[tb]
\centering
\caption{Training time comparison on the three datasets}
\label{tab:time}
\begin{tabular}{|l|c|c|c|}
\hline
\rule{0pt}{2.5ex} \textbf{Model} & \textbf{OpenImages} & \textbf{NUS-WIDE} & \textbf{IAPR-TC12}\\[0.5ex] \hline
VSE-ens &7.1h  &24.83h   &0.95h\\
WARP &35.65h      &51.46h      &2.38h\\
Opt-AUC  &10.13h    &25.06h   &1.82h\\
\hline
\end{tabular}
\end{table}

In Table~\ref{tab:complexity}, our model precomputes rankings every $|A|log|A|$ SGD update (as described in Algorithm~\ref{algo:sampling}), which can be finished in amortized runtime. Then it will draw a rank \emph{r}, the rank of negative sample in $O(1)$ and a latent factor \emph{f} in $O(k)$, resulting in the additional time complexity around $O(k)$. For the WARP model,  most time is consumed and determined by the negative sampling, which can be noted as $O(Tk)$. For Opt-AUC, although the time complexity for each SGD update is lowest among these models, it takes more training iterations for convergence since most negative examples selected by the uniform sampler are not informative.

In Table~\ref{tab:time}, our VSE-ens spends 7.1 hours in training on the OpenImages dataset, whereas WARP costs 5 times more training time. On the NUS-WIDE dataset, the improvement our model reaches is about 2x faster than WARP. Similar observation can be made on the IAPR-TC12 dataset. Besides, our proposed sampling also consistently takes shorter time than Opt-AUC, because VSE-ens requires less number of iterations to reach convergence. More specifically, our approach can reach the stable status and converge at around 200 iterations, WARP costs 150 iterations (thus more costly for each iteration), and Opt-AUC takes around 800 iterations to complete the optimization in our experiments.
\subsection{Comparison in Ranking Accuracy}
The ranking accuracy of all the comparison models is shown in Table~\ref{tab:accuracy}, where the percentage of improvements that our approach gains relative to WARP is also presented in the last row of each dataset. In general, our model achieves the best performance in ranking accuracy. Specifically, WARP is a stronger baseline than Opt-AUC, given the fact that the higher ranking accuracy is achieved across all the datasets. Our VSE-ens model outperforms WARP in all testing datasets, with a large portion of improvements. In particular, the improvements on NUS-WIDE are the most significant, which can reach up to around 166\% in terms of MAP. This implies that our adaptive negative samplers are more effective than the uniform samplers used by WARP and Opt-AUC. Note that the amount of improvements vary quite different among datasets, which may be due to the different statistics of our datasets, and require further study as part of our future work. 

In conclusion, our VSE-ens approach cannot only greatly reduce the training time in sampling positive-negative annotation pairs for each image, but also effectively improve the performance of image annotation in comparison with other counterparts across a number of real datasets. 

\section{Related Work}
%
Many approaches have been proposed in the literature to resolve the issue of semantic gap in the task of image annotation. In general, these approaches can be roughly classified into three types, namely (1) manual annotation, (2) semi-automatic annotation and (3) automatic annotation. Manual annotation requires users to provide the browsed images with descriptive keywords, which are often regarded as the ground truth of corresponding datasets. However, man power is often very expensive and it would be even intractable when facing a huge amount of images.   

Semi-automatic annotations can produce automatic annotation to some extent, but also require to build fundamental structures with the involvement of human beings. For example, \cite{marques2003semi} propose a layered structure to build image ontology for annotations, where low-level features of images are selected by the bottom layer. By abstracting low-level features up to high-level features, it connects the semantic feature of images with appropriate annotations. However, the building of image ontology requires expert knowledge, and may be domain-specific. 
\cite{zhang2010semi} formulate image annotation as a multi-label learning problem, and develop a semi-automatic annotating system. For a given image, their system initially chooses some keywords from a vocabulary as labels, and then refines these labels in the light of user feedback.  

Most existing works follow the direction of automatic image annotation, which provides the greatest flexibility and the least involvement of human users. To this end, some researchers make use of textual information for image annotation. \cite{deschacht2007text} present a novel approach to annotate images by the associated text. It first determines the salient and attractive parts of text from which semantic entities (e.g, persons and objects) are then extracted and classified. \cite{verma2012image} propose a two-step variant of K-nearest neighbor approach, where the first step is to learn image-to-label similarities and the second is to learn image-to-image similarities. Both kinds of similarities are combined together to help annotate an image with proper labels. \cite{uricchio2017automatic} propose a label propagation framework based on Kernel Canonical correlation analysis. It builds a latent semantic space where correlations of visual and textual features are well preserved. 

For visual semantic embeddings, \cite{frome2013devise} develop a new deep visual-semantic embedding model which transfers the semantic knowledge learned from a textual domain to a deep neural network trained for visual object recognition.
\cite{yu2013neighborhood} propose a multi-label classification method for automatic image annotation. It takes into consideration the uncertainty to map visual feature space to semantic concept space based on neighborhood rough sets. The label set of a given image is determined by maximum a posteriori (MAP) principles.  
\cite{DBLP:journals/corr/RenJLFY15} introduce a multi-instance visual-semantic embedding model to embed images with a single or multiple labels. This approach first constructs the image subregion set, and then builds the region-to-label correspondence. \cite{DBLP:journals/corr/KirosSZ14} describe a framework of encoder-decoder models to address the problem of image caption generation. The encoder learns a joint image-sentence embedding using long short-term memory (LSTM) and the decoder generates novel descriptions from scratch by a new neural language model. 

Different from the above works, our problem settings do not have associated text or content to describe images. Besides, our main focus is not to better model images, but to provide a better solution to find appropriate annotation pairs in shorter time, which may be beneficial for other models. 

\section{Conclusions}
In this paper, we aimed to resolve the problem of slow negative sampling for visual-semantic embeddings. Specifically, we proposed an adaptive sampler to select highly ranked negative annotations by adopting a rank-invariant transformation, through which the time complexity can be greatly reduced. We showed that our proposed sampling was theoretically comparable with traditional negative sampling based on time-consuming inner products. Experimental results demonstrated that our approach outperformed other counterparts both in training time and ranking accuracy. 


\section{Acknowledgments}
This work was supported by the National Natural Science Foundation for Young Scientists of China under Grant No. 61702084 and the Fundamental Research Funds for the Central Universities under Grant No.N161704001. We would like to thank Fartash Faghri for his insightful suggestions about the visual semantic embeddings.

\bibliography{VSE}
\bibliographystyle{aaai}
\end{document}